\documentclass[journal,twocolumn]{IEEEtran}
\usepackage[numbers,sort&compress]{natbib}
\usepackage[ruled]{algorithm2e}  
\usepackage{enumerate}
\usepackage{ulem}
\usepackage{multirow}
\usepackage{colortbl}
\usepackage{color}
\usepackage{array}
\definecolor{mygray}{gray}{.6}
\usepackage{graphicx}
\usepackage{subfigure}
\usepackage{float}  
\usepackage[table]{xcolor}
\usepackage{amsthm,amsmath,amssymb}
\usepackage{threeparttable}
\usepackage{mathrsfs}
\usepackage{graphicx}
\usepackage{booktabs} 
\usepackage{tabu}
\newcolumntype{C}[1]{>{\centering}p{#1}}
\usepackage{makecell}
\setcellgapes{4pt}
\usepackage{algorithmic}

\usepackage{float}
\usepackage{graphicx}

\begin{document}
	\title{Multi-objective optimization based network control principles for identifying personalized drug targets with cancer}
		\author{Jing Liang, Zhuo Hu, Zong-Wei Li, Kang-Jia Qiao, Wei-Feng Guo
		\thanks{This paper was supported by the National Natural Science Foundation of China (62002329,61922072,61876169),and National Ministry of Science and Technology key research and development (2022YFD2001200), and Key scientific and technological projects of Henan Province (212102310083), and China postdoctoral foundation (2021M692915), and Henan postdoctoral foundation (202002021), Research start-up funds for top doctors in Zhengzhou University (32211739), and open Funds of State Key Laboratory of Oncology in South China (HN2021-01). (\textit{Corresponding author:} Wei-Feng Guo.)
			
			Jing Liang, Zhuo Hu, Zong-Wei Li, Kang-Jia Qiao, Wei-Feng Guo are with School of Electrical and Information Engineering, Zhengzhou University, Zhengzhou 450001, China and State Key Laboratory of Intelligent Agricultural Power Equipment, Zhengzhou University, Luoyang 471100, China  (e-mail: liangjing@zzu.edu.cn; 1344051630@qq.com; 1032173185@qq.com; qiaokangjia@yeah.net; guowf@zzu.edu.cn).
	}}
	\markboth{Journal of \LaTeX\ Class Files,~Vol.~, No.~, ~}%
	{How to Use the IEEEtran \LaTeX \ Templates}
\maketitle

\begin{abstract}
It is a big challenge to develop efficient models for identifying personalized drug targets (PDTs) from high-dimensional personalized genomic profile of individual patients. Recent structural network control principles have introduced a new approach to discover PDTs by selecting an optimal set of driver genes in personalized gene interaction network (PGIN). However, most of current methods only focus on controlling the system through a minimum driver-node set and ignore the existence of multiple candidate driver-node sets for therapeutic drug target identification in PGIN. Therefore, this paper proposed multi-objective optimization-based structural network control principles (MONCP) by considering minimum driver nodes and maximum prior-known drug-target information. To solve MONCP, a discrete multi-objective optimization problem is formulated with many constrained variables, and a novel evolutionary optimization model called LSCV-MCEA was developed by adapting a multi-tasking framework and a rankings-based fitness function method. With genomics data of patients with breast or lung cancer from The Cancer Genome Atlas database, the effectiveness of LSCV-MCEA was validated. The experimental results indicated that compared with other advanced methods, LSCV-MCEA can more effectively identify PDTs with the highest Area Under the Curve score for predicting clinically annotated combinatorial drugs. Meanwhile, LSCV-MCEA can more effectively solve MONCP than other evolutionary optimization methods in terms of algorithm convergence and diversity. Particularly, LSCV-MCEA can efficiently detect disease signals for individual patients with BRCA cancer. The study results show that multi-objective optimization can solve structural network control principles effectively and offer a new perspective for understanding tumor heterogeneity in cancer precision medicine. 
The source code of our LSCV-MCEA and supplementary files can be freely downloaded from https://github.com/WilfongGuo/MONCP, with all data in this study.

\end{abstract}


\section{Introduction}
\IEEEPARstart{A}{dvances} in next-generation sequencing technologies have promoted the development of personalized genomic profiles of individual patients for large-scale target discovery in cancer \cite{kalari2018panoply,guo2021network,guo2021performance,liany2021druid}. In the past decades, the research on developing computational methods has begun to shift from drug target identification for population patients \cite{zheng2020predicting} to personalized drug target (PDT) recommendation based on the personalized genomic profile of a patient \cite{guo2021network,guo2021performance}. However, because of the limited number of the personalized samples information (e.g., the personalized omics data), it is difficult to select/find the effective individual patient-specific drug targets with personalized omics data in clinical for recommending personalized drug combinations.

With the development of network science, many sample-specific gene interaction network-based tools including CPGD \cite{guo2021network}, PNC \cite{guo2019novel}, SCS \cite{guo2018discovering}, and DawnRank \cite{hou2014dawnrank} could be used on personalized gene interaction networks (PGIN) to identify PDTs by selectively targeting personalized oncogenic drivers on PGIN. From a network perspective, an individual patient is generally a nonlinear dynamical system where gene expressions are variables of such a system and may be different if measured at different time points or conditions even for the same individual patient \cite{yazar2022single,reiter2018minimal}. In contrast, it is gene associations or transcriptional networks that result in the measured gene expression patterns, and thus is a stable form against the time and condition \cite{liu2016personalized,zhao2016part,van2018integrative,smart2023emergent}. Therefore, the network of an individual patient can more reliably characterize the biological system or state of the individual patient. Therefore, it is rational and effective in practical applications for identifying personalized drug targets from personalized gene interaction networks.
Among these methods, structural network control principles are popular for identifying PDTs \cite{guo2020network}, which can provide a theoretically accurate description of how the state transition can be achieved by a proper set of driver genes in PGIN. From the perspective of the structural network control principle, drug targets could be considered as targeting personalized driver genes (i.e., PDGs) on PGIN. These PDGs are candidate drug targets by the drug stimulation signals whose state transitions can change the whole network state from the initial state to the desired state \cite{guo2020network,guo2021network,guo2021performance}. Note that Paired-SSN \cite{guo2019novel} is used in this study to construct the PGIN by integrating their gene expression data of paired samples for an individual patient and gene mutation data of individual patients (\textbf{Figure 1A}). 

\begin{figure*}[!t]
	\centering
	\includegraphics[width=7.5 in]{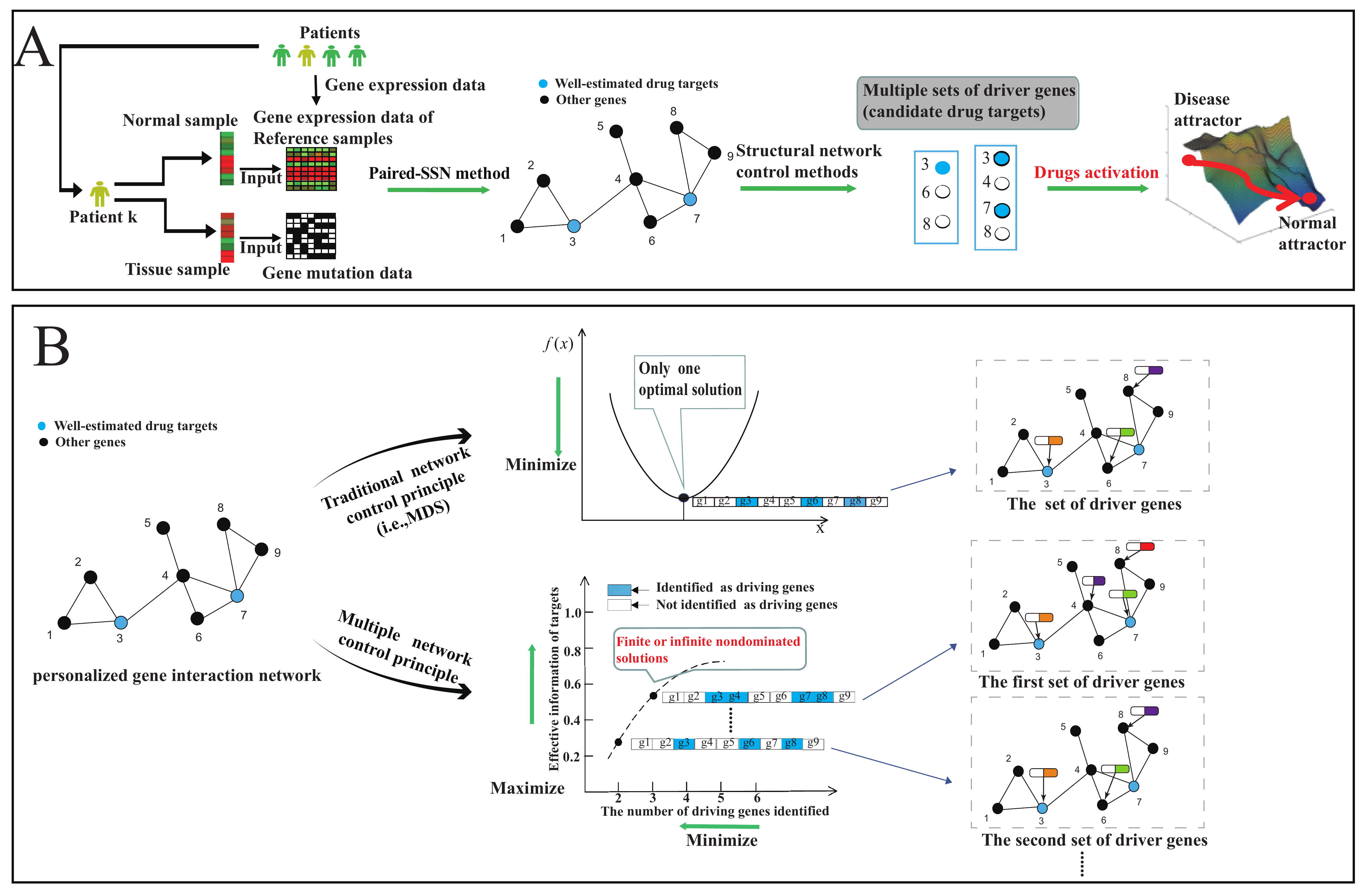}
	\caption{\textbf{Demonstration of our MONCP and framework of LSCV-MCEA.} (A) The problem formation of MONCP. We formed MONCP, which aims to discover multiple candidate driver node sets (MCDS) on the PGIN. The PGIN was constructed with Paired-SSN  by integrating their gene expression data of paired samples of an individual patient and gene mutation data. The selection of the MCDS should guarantee that the variables are constrained to make the whole network controllable for driving the personalized state transition between the normal state and disease state on the PGIN. (B) The demonstration of LSCV-MCEA. In fact,  current several popular network control methods including the DFVS, MDS and NCUA can be calculated using Integer Linear Programming  formalization. This formalization could be considered as  a discrete optimization problem with large-scale constrained variables. In this study, we considered our MONCP as a DMOM-LSCV and proposed a novel algorithm called LSCV-MCEA for providing the PF to discover MCDS of MONCP. First, the PGIN was divided into many subnetworks which form the initial multi-population (via population initialization strategy). Then, we optimized the gene selection of the MCDS by using the evolutionary multitasking strategy and the rankings-based fitness function strategy on the population. Finally, after reaching the termination condition on PGIN, the LSCV-MCEA outputted MCDS (i.e., non-dominated solutions of DMOMs-LSCV) rather than one minimum driver node set of traditional structural network control principles.}
	\label{fig1}
\end{figure*}  

In fact, achieving control over a PGIN from a set of PDTs is essentially an optimization problem to select optimal PDGs for driving the state transition between the disease state and the normal state through drug stimulation \cite{guo2020network,guo2021network,guo2021performance}. The PDGs that are targeted by efficient drugs are called targeting PDGs and are regarded as PDTs. The constraints for the variables are to guarantee that the network is controllable for the state transition. This significantly advances the structural control principles of complex networks in the understanding of the problem of controlling PGIN with complex dynamics, such as the Maximum Matching Sets (MMS)-based control method \cite{liu2011controllability}, the Directed Feedback Vertex-based control method (DFVS) \cite{zanudo2017structure}, the Minimum Dominating Sets (MDS)-based control method \cite{nacher2012dominating}, and the Nonlinear Control of undirected networks Algorithm (NCUA) \cite{guo2019novel}. However, these methods based on structural network control principles have two main limitations. One is that these methods are computationally difficult to find the optimal set of driver nodes in large-scale networks with nonlinear dynamics, and approximate efficient solutions are achievable \cite{guo2020network}. The other limitation is that these methods only focus on controlling the system through a minimum driver-node set, which is a single objective optimization-based control and ignores the prior drug targets for identifying optimal driver genes. 

Based on the above considerations, this paper first proposed the concept of \uline{m}ulti-\uline{o}bjective optimization-based structural \uline{n}etwork \uline{c}ontrol \uline{p}rinciples (MONCP) under the framework of DFVS, MDS, and NCUA in the PGIN. MONCP not only requires controlling the system with minimum driver nodes but also needs to find maximum prior-known drug-target genes in the PGIN (\textbf{Figure 1A}). In multi-objective optimization, a solution cannot optimize all objectives simultaneously due to the contradiction among different objectives. Multi-objective optimization aims to find a set of best trade-off solutions called the Pareto optimal solution Set (PS), and the corresponding objective values to the PS are called the Pareto Front (PF) \cite{qiao2022feature}. Different from the traditional structural network control principles that only provide driver nodes with a minimum number, MONCP can provide multiple sets of driver nodes for individual patients and provide solutions with more drug target information, thus offering more possible ways of personalized therapy in precision medicine.

Then, this paper converted MONCP to a \uline{d}iscrete \uline{m}ulti-objective \uline{o}ptimization \uline{m}odel with \uline{l}arge \uline{s}cale \uline{c}onstrained \uline{v}ariables (DMOM-LSCV) to seek optimal solutions of MONCP. DMOM-LSCV is a type of constrained multi-objective optimization problem (CMOP), where the variables are discrete and constrained in the decision space. In the past two decades, constrained multi-objective evolutionary algorithms (CMOEAs) are powerful tools for addressing the CMOPs \cite{liang2022survey}. Among these algorithms, the evolutionary multi-tasking strategy is a promising pattern for solving CMOPs owing to its powerful search capability and good scalability during the evolution process. The large-scale variables of our DMOM-LSCV make search space large and there are lots of constraints in decision space.
However, the current CMOEAs lacked the proper techniques for processing the large-scale constraints on high-dimensional decision variables, which may reduce the performance for seeking optimal solutions for DMOM-LSCV \cite{li2021large}. Thus, novel strategies must be designed to solve DMOM-LSCV.

To solve this DMOM-LSCV, this paper developed a \uline{l}arge-\uline{s}cale \uline{c}onstrained \uline{v}ariables based \uline{m}ulti-tasking \uline{c}ooperative \uline{e}volutionary algorithm (LSCV-MCEA) (\textbf{Figure 1B}). The proposed LSCV-MCEA adopts a multi-tasking framework that involves the main task to optimize the original CMOP with two objectives to approach PF, and auxiliary tasks to optimize the CSOP with each of two objectives. In short, the main population is mainly targeted at global search of PF, while the auxiliary populations focus on local search to help the main population explore the undeveloped area of PF. Additionally, a rankings-based fitness function was introduced into the auxiliary task to help the evolution of the main task \cite{liang2021differential}. To our best knowledge, LSCV-MCEA is the first method to introduce a multi-objective optimization mechanism into the evolutionary algorithm with the network control principle in network science.

Finally, the significant improvement of LSCV-MCEA in identifying PDTs was investigated on three cancer datasets (i.e., breast invasive carcinoma (BRCA), lung adenocarcinoma (LUAD), and lung squamous cell carcinoma (LUSC)) from The Cancer Genome Atlas (TCGA). The main contributions of this paper are listed as follows: 

\begin{list}{}{}
	
	\item{i) A novel concept of multi-objective optimization based on the structural network control principles (MONCP) is proposed under the framework of DFVS, MDS, and NCUA for drug targets identification in the PGIN. Compared with the traditional structural network control principles, MONCP can provide solutions with more drug target information for personalized therapy in precision medicine. }
		
	\item{ii) Based on MONCP, a discrete multi-objective optimization model with many constrained variables (i.e., DMOM-LSCV) is developed. Furthermore, we validated that the LSCV-MCEA can more efficiently optimize the DMOM-LSCV in terms of algorithm convergence and diversity than other state-of-the-art CMOEAs.}
	
	\item{iii) Also, it was validated that the evolutionary strategies (i.e., the multitasking strategy and the rankings-based fitness function strategy) can effectively improve the performance of the LSCV-MCEA including the precision for discovering the clinical efficacious combinatorial drugs and algorithm convergence and diversity for solving MONCP.}
	
	\item{iv) The PF found by our LSCV-MCEA is significantly enriched in the Cancer Gene Census (CGC) dataset. Particularly, the PF of our LSCV-MCEA could effectively detect the disease signals of stage iiib for individual patients with BRCA cancer.}
\end{list}

The remainder of this paper is organized as follows. Section II introduces the background knowledge of datasets, structural network control, and constrained multi-objective optimization. Section III presents the related work about structural network control methods and CMOEAs. Section IV provides multi-objective optimization-based structural network control principles (MONCP) and the design of CMOEA. Section V presents the experimental results on the three cancer datasets from TCGA. Section VI concludes this paper.

\section{PRELIMINARIES}
 This section introduces the background knowledge of datasets, structural network control, and constrained multi-objective optimization. 
\subsection{DataSets}

\subsubsection {The PGIN data constructed by Paired-SSN on the cancer gene omics data} 
Considering that Paired-SSN can more accurately characterize the personalized state transition between the normal state and the disease state compared with other SSN methods \cite{zhang2022prioritization}, this paper constructed the PGIN with Paired-SSN \cite{guo2019novel} by integrating their gene expression data of paired samples of an individual patient and gene mutation data of individual patients (\textbf{Figure 1A}). In this study, gene expression datasets of the breast invasive carcinoma patients (BRCA), lung adenocarcinoma patients (LUAD), and lung squamous cell carcinoma patients (LUSC) from the Cancer Genome Atlas (TCGA) were used to construct PGIN and make comparisons. Normal samples and tumor samples were extracted from the same patient, and they were denoted as paired samples of individual patients and obtained from the TCGA data portal \cite{guo2021network}. The number of patients with BRCA was 112, and there were 49 LUSC patients and 57 LUAD patients, respectively. Note that when constructing the PGIN for LUSC and LUAD, reference samples were chosen from the 106 normal samples of the patients with lung cancer.

To form the reliable edges in PGIN, in addition to the three gene expression datasets (i.e., BRCA, LUSC, and LUAD) from TCGA, the single nucleotide variations were used in this paper, which contained 90490, 72541, and 65304 non-synonymous somatic mutations, for the SNV data of BRCA, LUAD, and LUSC datasets from TCGA, respectively. As results, for BRCA, LUSC, and LUAD patients the number of genes when solving the problem in PGIN range from 1581 to 1780, 1987 to 2342 and 2100 to 2344, respectively.  The specific calculation method is shown in Section S-I in the additional file 1.

\subsubsection  {Combinatorial drug and gene interaction network}
To evaluate the effectiveness of the drug targets of different methods in clinical, this paper collected combinatorial drug and gene interaction networks to prioritize the potential personalized anti-disease combinatorial drugs by measuring the number of targeting drug targets for a given drug combination. The interactions between combinatorial drugs and genes were collected by Quan et al. \cite{quan2018facilitating}. Such a combinatorial drug-gene interaction network contains 342 combinatorial drugs, among which 122 are efficacious for treating cancer. Especially, 32 and 17 prior-known clinical annotated combinatorial drugs (CAC) derived from DCDB are efficacious in treating breast cancer and lung cancer, respectively.

\subsection{Network control in cancer}
Cancer is considered as a dysfunction of molecular networks, in which dynamics and changeability are the main characteristics of the molecular networks
\cite{wiener2019cybernetics,guo2020network}. Particularly, this paper considered the cancer progression from a normal state to a disease state as a network control problem. Regarding the gene expression profile in normal and tumor samples as the respective state of a given patient, this paper aimed to detect a small number of personalized driver genes (i.e., PDGs) by considering the structure of PGIN. These PDGs are candidate drug targets by drug activation signals, and they are controllers making this transition from the normal state to the disease state for the PGIN of each individual. It can be summarized as follows \cite{sun2017closed} :
\begin{equation}
	\frac{{{\rm{d}}x}}{{{\rm{d}}t}} = S(x,A) + Bu
\end{equation}
where $x \in {R^n}$ represents the state of genes at time $t$ in a PGIN; $n$ is the number of genes in PGIN. $S$ is the function in terms of the enhancement of the activity of the nodes in PGIN. $A \in {R^{n \times n}}$ denotes the state transition matrix (i.e., adjacency matrix) of PGIN; $B \in {R^{n \times {n_c}}}_{}$ is ${n_c}$ controllers that can produce the input signals to make the state transition of the whole network with $n$ variables/nodes. The controllers in structural network control models denote the drug activations that can produce input signals to make the state transition of the whole network. In many cases, it is assumed that one driver node can be altered by an independent input signal \cite{vierbuchen2010direct,ieda2010direct}. The element to ${B_{{\rm{ij}}}}$ is non-zero if the $j-th$ input signal directly acts on node $v_i$. The input matrix is set as an identity matrix by considering all the network nodes as candidate control nodes. In this case, drug target identification is to find the feasible subset of genes $K$ from the candidate control genes set $U$, which are injected by properly inputting drug activation signals to a complex nonlinear individual system in cancer from the disease attractor (state) to the normal attractor (state) while satisfying the following requirements:
\begin{equation}
	u({v_i}) = \left\{ \begin{array}{l}
		{u_i},{\rm{if}}\;{v_i} \in K\\
		0,{\rm{otherwise}}
	\end{array} \right.
\end{equation}

\subsection{Constrained multi-objective optimization}
To form the MONCP, a Constrained Multi-objective Optimization Problem (CMOP) is formulated as:
\begin{equation}
	{\rm{min\;F(x)}} = ({f_1}(x),{f_2}(x),...{f_m}(x))
\end{equation}
\begin{equation}
	{\rm s.t.}\;\left\{ \begin{array}{l}
		x \in \Omega \\
		{g_i}(x) \le 0,i = 1,...,k\\
		{h_j}(x) = 0,j = 1,...,p
	\end{array} \right.
\end{equation}
where $\Omega$ is the decision space, and $x$ is a decision variable (i.e., the gene state of PGIN); $F$ represents $m$ objective functions to be optimized; ${g_i}(x)$ represent $k$ inequality constraints, and ${h_j}(x)$ represent $p$ equality constraints. Note that the decision vector can be continuous or discrete. In this paper, discrete decision variables are used to indicate whether the drug target is recognized, where 0 means that the drug target is not identified, and 1 means that it is recognized. Meanwhile, there are only inequality constraints in the decision space, which maintain the controllability of the network.  If the constraints are satisfied, the solution is feasible (the network is controllable); otherwise, it is infeasible (the network is uncontrollable). 

For two feasible solutions ${x_1}$ and ${x_2}$, if ${f_i}({x_2}) \ge {f_i}({x_1})$ for every $i \in \left\{ {1,...,m\} } \right.$ and ${f_j}({x_2}) > {f_j}({x_1})$ for at least one $j \in \left\{ {1,...,m\} } \right.$, ${x_1}$ is said to dominate ${x_2}$. If a solution is not dominated by any solution, it is called Pareto optimal solution. All Pareto optimal solutions constitute a Pareto optimal solution set (PS), and the mapping of PS in the objective space is PF. 

\section{Related Work}
In this section, the related work of structural network control principles and constrained multi-objective evolutionary algorithms are mainly introduced.

\subsection{Structural network control principles}

The control process of PGIN is usually determined by an intrinsic structure and dynamic propagation \cite{guo2020network}. Although the underlying wiring diagram in PGIN is known, the knowledge of the specific functional forms (i.e., the form of function $S$) of PGIN is lacking. Analyzing such complicated systems requires the concepts and approaches of structural network control principles, which can be exploited to identify driver nodes to drive the state transition between the normal state and the disease state for an individual patient. The driver nodes targeted by efficient drug activation can be regarded as candidate drug targets for drug discovery. The current structural network control principles consist of the Maximum Matching Set (MMS)-based control methods \cite{liu2011controllability}, Directed Feedback Vertex-based control method (DFVS) \cite{zanudo2017structure}, Minimum Dominating Set-based control method (MDS) \cite{nacher2012dominating}, and Nonlinear Control of undirected network Algorithm (NCUA) \cite{guo2019novel}. Among them, MMS-based methods could solve the control problem of large-scale networks in polynomial time, but they assume unspecified linear dynamics in complex networks, which may only give an incomplete view of the network control properties of PGIN with nonlinear dynamics. The details of DFVS, MDS, and NCUA for the PGIN will be discussed below.

For MDS-based control methods, the driver nodes can be identified by assuming that each edge in PGIN is bidirectional \cite{nacher2012dominating}. The identification of MDS can be solved by using Integer Linear Programming (ILP) formalization as follows:

\begin{equation}
	\begin{array}{l}
		\min \begin{array}{*{20}{c}}
			{\sum\limits_{i \in V} {{x_i}} }&{}
		\end{array}\\
	\end{array}
\end{equation}
\begin{equation}
	{\rm s.t.}\;{x_i} + \sum\limits_{j \in \partial (i)} {{x_j}}\\  \ge 1,{x_i},{x_j} \in \{ 0,1\} 
\end{equation}
where $\partial(i)$ denotes the neighborhood nodes of node $i$; $x_i$ will take the value 1 when node $i$ belongs to the MDS.

By considering the PGIN as a directed graph, DFVS illustrates that the driver nodes can be determined by the cycle structure and the source nodes of a directed network with the Feedback Vertex Sets (FVS). The FVS can be calculated by using ILP formalization \cite{balakrishnan1995exact}, which adopts a scheme that assigns weights to vertices to capture an ordering relationship among the vertices. The ILP is formalized as follows:
\begin{equation}
	\begin{array}{l}
		\min \begin{array}{*{20}{c}}
			{\sum\limits_{i \in V} {{x_i}} }
		\end{array}\\
		
	\end{array}
\end{equation}
\begin{equation}
	\begin{split}
		{\rm s.t.}\;{w_i} - {w_j} + n{x_i} \ge 1({\rm every}\;{v_i} \to {v_j}) \\
		0 \le {w_i} \le 1,{x_i},{x_j} \in \{ 0,1\}. 
	\end{split}	
\end{equation}
where the variable $x_i$ will take the value 1 when node $i$ belongs to the FVS; Note that the source nodes could be directly obtained by using the structure of PGIN with edge directions.

NCUA considers the PGIN as an undirected network $G (V, E)$, and it assumes that each edge in PGIN is bidirectional. For NCUA, $G (V, E)$ is converted into a bipartite graph $G (V_T, {{V_ \bot }},E_1)$, where ${{V_T} \equiv {V }}$ and $V_ \bot \equiv {E}$. If $v_i  \in V_T$ is one of the nodes for $v_j\in {{V_ \bot }}$, an edge connecting $v_i$ and $v_j$ is added into the set $E_1$. After the bipartite graph is obtained, a modified version of the dominating set is adopted. The dominating set must be selected from $V_T$, and it is also sufficient to dominate all of the nodes in  ${{V_ \bot }}$. The minimum dominating set cover problem can be solved by the following ILP model:

\begin{equation}
	\begin{array}{l}
		\min \begin{array}{*{20}{c}}
			{\sum\limits_{v \in {V_T }\equiv {V}} {{x_v}} }&{}
		\end{array}\\
		
	\end{array}
\end{equation}
\begin{equation}
	{\rm s.t.}\;\sum\limits_{\{ v,u\}  \in {E_1}} {{x_v}}  \ge 1({\rm every}\; u \in {V_T} \equiv {V}),{x_v} \in \{ 0,1\} 		
\end{equation}
where the variable $x_i$ will take the value 1 when node $i$ belongs to the cover set in ${V_T}$ to cover the nodes in  ${{V_ \bot }}$.

For a given network, there are a large number of variables and constraints for the optimization of MDS, DFVS, and NCUA. It is well known that the optimization of MDS, DFVS, and NCUA is an NP-hard problem, but approximate efficient solutions are still achievable \cite{guo2020network}. Meanwhile, they only focus on controlling the system through any minimum driver-node set but ignore the existence of multiple candidate driver-node sets (MCDS) for drug target identification in PGIN. To address this issue, this paper proposed the concept of multi-objective optimization based structural network control principles (MONCP) under the framework of DFVS, MDS, and NCUA in the PGIN and formed an Evolutionary Constrained Multi-objective Optimization model (DMOM-LSCV) to improve the performance for seeking optimal solutions to MONCP.

\subsection{Existing constrained multi-objective evolutionary algorithms}

In the past few years, many variants of CMOEAs have been proposed to solve CMOPs. In these methods, constraint-handling techniques (CHTs) and optimizers are important components for addressing CMOPs \cite{qiao2022evolutionary}. For example, Deb et al. proposed NSGA-II-CDP by integrating the constrained dominance principles (CDP) into the NSGA-II optimization framework to obtain feasible non-dominated solutions while satisfying the constraints \cite{deb2002fast}. Recently, by integrating $\theta {\rm{-}}dominance$ and CDP to form a multi-ranking strategy, Ming et al. \cite{ming2022constrained} proposed a constrained multi-objective evolutionary algorithm called CMME with enhanced mating and environmental selection, thus achieving a balance of different goals.
However, these methods are easy to fall into the local optimum when complex constraints distort the search space and lead to many discontinuous feasible regions, increasing the difficulty of solving CMOPs. 

With the development of multi-tasking learning \cite{gupta2015multifactorial}, Evolutionary multitasking (EMT) has become a promising tool to solve CMOPs with complex constraints by utilizing the relationships between different optimization tasks. The main idea of these methods is described as follows: First, the problem is transformed into a multi-task optimization problem, in which the auxiliary task assists in solving the main task. Multi-population is a representative optimization form: the population is divided into one main population and multi-auxiliary populations; then the evolution of these populations is considered as multiple distinct but related optimization tasks; finally, the search ability and convergence performance of the algorithm on each task will be improved by sharing the knowledge of different tasks during the evolution. After several years of development, many multi-tasking optimization methods have been proposed, such as MTCMO \cite{qiao2022dynamic}. Also, CCMO \cite{tian2020coevolutionary}, c-DPEA \cite{ming2021dual}, and  CCMODE \cite{wang2018cooperative} have similar evolutionary forms with multi-tasking algorithms, so they can be regarded as multi-tasking algorithms.

However, for DMOM-LSCV, the large-scale constraints make the constrained landscape in the decision space different and much more complex compared with test problems of current CMOEAs. It is easy to fall into local optimum and difficult to find feasible solutions for existing CMOEAs facing with large-scale constraints in decision space. To solve this problem, this paper proposed a novel CMOEA denoted as LSCV-MCEA to solve MONCP, and detailed information on LSCV-MCEA is presented in the next section.

\begin{figure*}[!t] 
	\centering
	\includegraphics[scale=0.75]{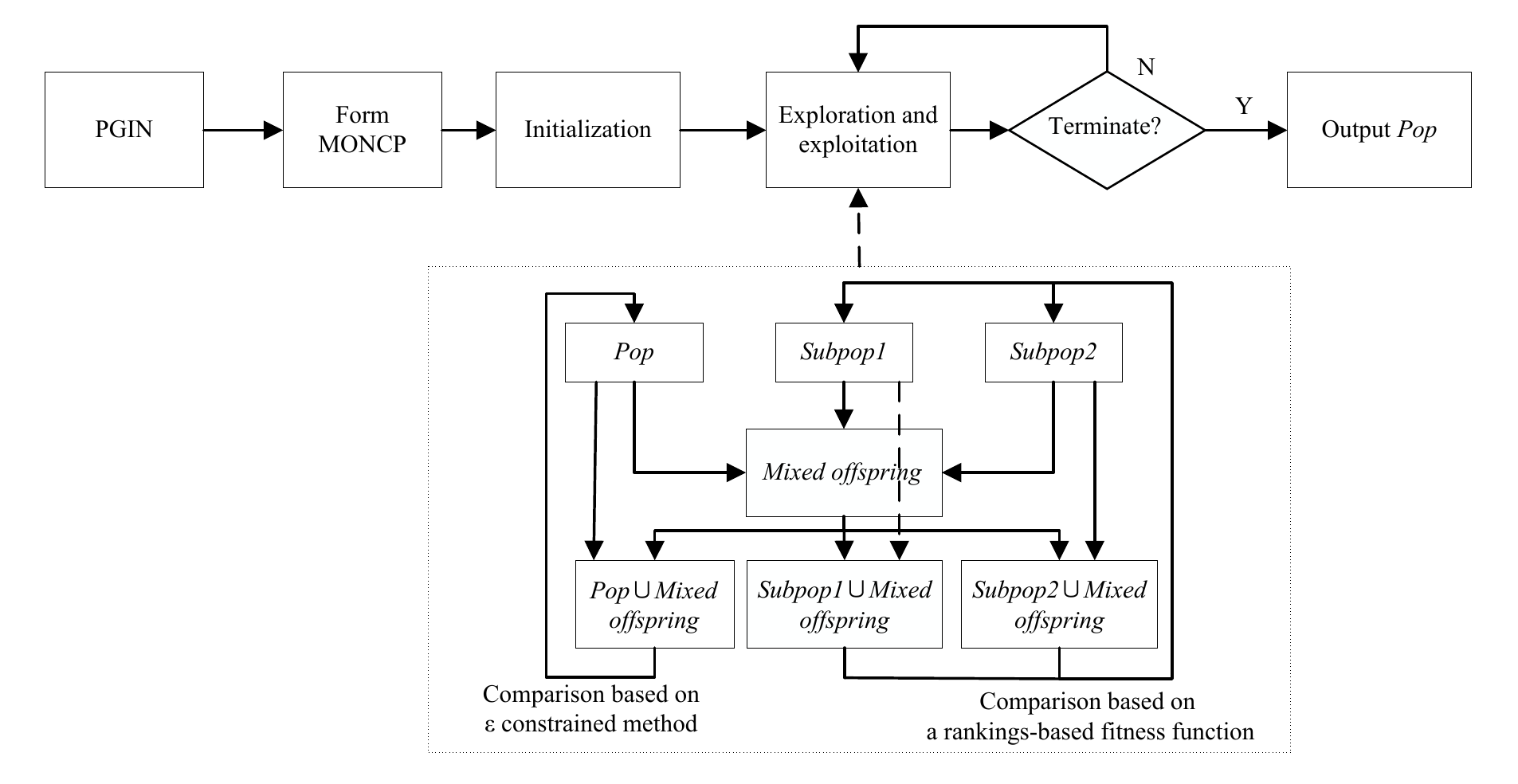}
	\caption{
		\textbf{Flowchart of LSCV-MCEA}. First, we form MONCP according to constructed PGIN. Then, a main population (denoted as $Pop$) with the task of all objectives and constraints and two subpopulations  (denoted as $Subpop1$ and $Subpop2$) with the task of corresponding objective and constraints are initialized. In each  generation, the three populations evolve independently to generate the offspring populations. All the offsprings are combined into a mixed offspring (denoted as $Mixed offspring$). This mixed offspring is combined to form one population which are integrated with $Pop$, $Subpop1$ and $Subpop2$, respectively. The three combined populations is used to select final excellent solutions by the environment selection. It is worth noting that the main population $Pop$ applys $\varepsilon$ constrained method to select final excellent solutions but two subpopulations  apply  rankings-based fitness function to select final excellent solutions. Finally, the main population $Pop$ is returned as the final output.}
	
	\label{fig2}
\end{figure*}

\section{THE PROPOSED MODEL}

\subsection{Problem formation of MONCP}
Under the framework of DFVS, NCUA, and MDS, the minimum driver nodes and maximum prior-known drug-target information are considered to control the system. Meanwhile, it is necessary to satisfy various constraints of nodes/variables to ensure that the network is controllable. Therefore, two objectives and constraints are used to identify PDTs in PGIN for MDS, DFVS, and NCUA. One objective function is minimum driver nodes, as shown in Equation (11); the other objective function is maximum drug targets among driver nodes as shown in Equation (12) according to the definition of MONCP. The fewer genes we selected (the first objective (i.e., equation (11)) is better), the less information of prior-known drug targets the selected genes would contain (the second objective (i.e., equation (12)) will be worse). Obviously, the two objectives of MONCP are conflicting. Therefore, for our MONCP multiple sets of non-dominated solutions need to be found. The constraints (i.e., the network is controllable) are defined in Equation (13).

	\begin{equation}
		min{\rm{ }}{f_1} = \left\| D \right\| = \sum\limits_{i \in V} {{x_i}} 
	\end{equation}
	\begin{equation}
		max{\rm{ }}{f_2} = \left\| {D \cap DT} \right\| = \sum\limits_{i \in V} {l{}_i{x_i}} 
	\end{equation}
	\begin{equation}
		{\rm s.t.}\; {g_k}(x) \le 0,k = 1,2...,m
\end{equation}
where $x = ({x_1},{x_2} \ldots {x_n})$ denotes the decision vector for the variables (or nodes) in the node set of network $V$ with binary encoding; $D$ and $DT$ represent the set of identified drug targets (i.e., targeting driver genes) and the set of prior-known drug-targets, respectively. If the gene is selected, ${x_{\rm{i}}}$=1; otherwise, ${x_{\rm{i}}}$ =0. $l = ({l_1},{l_2} \ldots {l_n})$ denotes the label vector with binary encoding. If the gene is within the set of prior known drug targets, ${l_{\rm{i}}}$=1; otherwise, ${l_{\rm{i}}}$ =0. $n$ represents the number of genes in PGIN, and $m$ denotes the number of constraints. Note that for MDS, DFVS, and NCUA, the objective functions are all the same, but the constraints could be expanded as Equations (6), (8), and (10) respectively. The constrains guarantee the network controllable. That is, all nodes could be controlled or regulated by the selected driver nodes. Different structural network control principles have different constraints under their corresponding assumptions. DFVS and NCUA illustrates that the driver nodes can be mainly determined by the feedback vertex sets (FVS) of a directed and undirected network respectively. The MDS assums that each edge is bidirectional for undirected network and the driver nodes could be selected from the minimum dominating sets.

The traditional MDS, DFVS, and NCUA with a single optimal objective (i.e., the minimum number of driver genes) can be approximately solved by using the LP-based classic branch and bound method \cite{lenstra1983integer}. However, this method ignores the existence of MCDS for solving our MONCP in the PGIN (i.e., Equations (5)-(6) for MDS, Equations (7)-(8) for DFVS and Equations (9)-(10)) for NCUA). That is, the traditional network control principles with a single objective can find approximate efficient solutions instead of a group of solutions to optimize the two objectives for solving MONCP.

\subsection{Motivation of LSCV-MCEA}
Our MONCP is a CMOP involving many discrete variables constrained in the decision space. In the past years, evolutionary multi-tasking frameworks with multiple populations have proven to be successful in solving large-scale CMOPs \cite{ge2017distributed}.  However, many of the current state-of-the-art CMOEAs struggle to achieve high performance when faced with large search space and numerous constraints in the decision space. Thus, this paper extended the existing evolutionary multi-tasking framework to solve MONCP. 

When solving MONCP with many discrete variables constrained in the decision space, the main obstacle is that there may be few or no feasible solutions in the early stage of the evolution. To address this issue, a rankings-based fitness function is applied to the sub-populations, which can help the main population to cross the infeasible area and maintain diversity.

\subsection{Creation of the auxiliary task}
In this paper, there are two auxiliary tasks:
\begin{equation}
	{\rm{min\;F(x)}} = {f_1}(x) \ {\rm or} \ {f_2}(x)
\end{equation}
\begin{equation}
	{\rm s.t.}\; {g_k}(x) \le 0,k = 1,2...,m
\end{equation}
where the first auxiliary task optimizes ${f_1}(x)$ and the other optimizes ${f_2}(x)$.

Different from the main task, the auxiliary tasks are CSOPs that can explore the undeveloped area of the main task from different directions and maintain the diversity of the main task. Meanwhile, the constraint boundary gradually reduces to help the main population cross the infeasible area. The auxiliary task is simpler, so it can help to solve the main task. Based on this, a new LSCV-MCEA framework is proposed and introduced below.

\subsection{Framework of LSCV-MCEA}
{\textbf{Figure 2} illustrates the framework of LSCV-MCEA. i) First, a main population (denoted as $Pop$) and two auxiliary populations (denoted as $Subpop1$ and $Subpop2$) are initialized randomly. The main population is used to optimize the original CMOP, which contains two objectives and all of the constraints; each auxiliary population is used for the corresponding task, which contains only one objective and all of the constraints. ii) Then, in each generation, the three populations evolve independently to generate the offspring populations. All the offspring populations are combined into a mixed offspring population (denoted as mixed offspring), which is integrated with $Pop$, $Subpop1$, and $Subpop2$, respectively. The sharing of offspring information is to ensure the diversity of these offspring populations \cite{qiao2022dynamic}. 
iii) Finally, three combined populations are integrated to form a new population that is used for the next generation update. Note that the main population $Pop$ uses the $\varepsilon$ constrained method to select the final excellent solutions, but the two sub-populations use a rankings-bassed fitness function to select the final excellent solutions until the termination condition is satisfied. Additionally, the rankings-based fitness function is employed to ensure the diversity of the main population, avoid premature convergence, and help the main population to cross the infeasible area \cite{liang2021differential}. Here, $Pop$ is returned as the final output.} The process is also provided in \textbf{Algorithm 1}. 

\begin{algorithm}[!h]
	\algsetup{linenosize=\tiny} \scriptsize
	\caption{Procedure of LSCV-MCEA }
	\label{alg:AOS}
	\renewcommand{\algorithmicrequire}{\textbf{Input:}}
	\renewcommand{\algorithmicensure}{\textbf{Output:}}	
	\begin{algorithmic}[1]
		\REQUIRE $PGIN$ (Network constructed)   
		\ENSURE The set of driver nodes for  individual patients   
		\STATE  Form MONCP according to the PGIN;
		\STATE  Initialize $Pop$ of size $N$ and $Subpop1$, $Subpop2$ of size $N_1$ randomly; // Pop is the main population, and $Subpop1$, $Subpop2$ are auxiliary populations.
		\STATE  Evaluate $Pop$ on the CMOP. // Evaluate the fitness for the solutions.
		\STATE  Evaluate $Subpop1$ and $Subpop2$ on the corresponding CSOP by the rankings-based fitness function ; 
				
		\WHILE{$Termination\ conditions \ are \ not \ satisfied\ $}				
		\STATE $MP_1$ $\leftarrow $ Select $N/2$ individuals as parents from $Pop$ by mating selection of MOEA; // Generate a mating pool from $Pop$.
		\STATE  $O_1$ $\leftarrow $ Generate $N/2$ offsprings based on $MP_1$ by the operators of MOEA; // Generate $N/2$ offspring.
		\STATE $MP_2$  $\leftarrow $Select $N_1/2$ individuals as parents from $Subpop1$ by mating selection of MOEA; 
		\STATE  $O_2$ $\leftarrow $ Generate $N_1/2$ offsprings based on $MP_2$ by the operators of MOEA  ;	
		\STATE  $MP_3$ $\leftarrow $ Select $N_1/2$ individuals as parents from $Subpop2$ by mating selection of MOEA; 
		\STATE  $O_3$ $\leftarrow $ Generate $N_1/2$ offsprings based on $MP_3$ by the operators of MOEA  ;	
		\STATE $Pop$ $\leftarrow$ $Pop$ $\cup $ $O_1$ $\cup$ $O_2$ $\cup$ $O_3$; // combine $Pop$ with the three offspring populations;
		\STATE $Subpop1$ $\leftarrow$ $Subpop1$ $\cup$ $O_1$ $\cup$ $O_2$ $\cup$ $O_3$; 
		\STATE $Subpop2$ $\leftarrow$ $Subpop2$ $\cup $ $O_1$ $\cup$ $O_2$ $\cup$ $O_3$; 

		\STATE  Evaluate $Pop$ on the CMOP ;
		\STATE   Evaluate $Subpop1$ and $Subpop2$ on the corresponding CSOP by the rankings-bassed fitness function;
		\STATE  $Pop$ $\leftarrow $ Select $N$ solutions from
		$Pop$ through $\varepsilon$ constrained method;
		\STATE  $Subpop1$ $\leftarrow $ Select $N_1$ solutions from
		$Subpop1$ by comparing fitness; 	
		\STATE  $Subpop2$ $\leftarrow $ Select $N_1$ solutions from
		$Subpop2$ by comparing fitness; 
		\ENDWHILE		
		\STATE A set of driver nodes $\leftarrow $ The non-dominated solutions of $Pop$.	
		\RETURN The set of driver nodes for  individual patients
	\end{algorithmic}
\end{algorithm}

\subsection{Computational complexity and calculation times}
The main time complexity comes from the following aspects in LSCV-MCEA at each generation: population initialization, fitness evaluation, crossover operator, mutation operator and environmental selection strategy for main population and auxiliary population respectively. The complexities of above aspects are $O({N_1}n)$, $O({T}{N_1}n^2)$, $O({T}{N_1}n)$, $O({T}{N_1})$ and $O({T}{N_1}^2)$ in the main population, respectively. In summary, the overall computational complexity in the main population is about $O({T}{N_1}n^2)$. In auxiliary population, the complexities of the above strategies are $O({N_2}n)$, $O({T}{N_2}n^2)$, $O({T}{N_2}n)$, $O({T}{N_2})$ and $O(G{N_2}^2)$, in which the whole computational complexity of auxiliary populations is approximately $O({T}{N_2}n^2)$. $N_1$ and $N_2$ represent the popsize of main population and auxiliary population. $T$ denotes  the number of iterations and $n$ denotes the number of genes in PGIN. To sum up, the total computational of LSCV-MCEA is $O({T}{N_1}n^2)$ on the PGIN for a individual patient.

Furthermore, we compared calculation time of different algorithms on three cancer datas. All the experiments used a specialized computer with Intel(R) Xeno(R) Gold 6230 CPU and 640 GB RAM. Please note that all experiments were implemented on PlatEMO \cite{tian2017platemo}. The results of the comparison are presented in Figure S6 in the additional file 1. Under the framework of MDS, NCUA and DFVS,  the calculation time of LSCV-MCEA is shorter than that of the other two CMOEAs (i.e., CCMO and MTCMO). Although the calculation time of single objective based network controllability algorithms (i.e., MDS, NCUA, and DFVS) is generally shorter than that of LSCV-MCEA, these methods for identifying one minimum set of driver nodes cannot effectively identify drug targets compared with LSCV-MCEA for discovering multiple sets of driver nodes.

\subsection{Code and data availability}
The code of our LSCV-MCEA can be freely downloaded from https://github.com/WilfongGuo/MONCP, with all data in this study. Supplementary files 1-7 in our study are available at https://github.com/WilfongGuo/MONCP/tree/main/Additional\\
Files.

\section{EXPERIMENT}
\subsection{Experimental settings}
In this work, the following experimental settings were adopted.

\subsubsection{Parameter settings of LSCV-MCEA}  
The parameters of LSCV-MCEA consist of the size of the main population, the size of the auxiliary populations, and the maximal number of function evaluations. Considering the variables of DMOM-LSCV, the size of the main population was generally set as $N=300$ for each PGIN \cite{he2020paired}. Meanwhile, the size of the auxiliary populations was set as $0.3*N=90$ \cite{wang2018cooperative}. Besides, the maximal number of function evaluations was set to 100,000 for each run \cite{qin2021large}.

\subsubsection{Algorithms for comparison} 
In this paper, LSCV-MCEA was compared with the other 19 methods, which are categorized into three groups below according to their significant features.

i) CMOEAs. MTCMO \cite{qiao2022dynamic}, CCMO \cite{tian2020coevolutionary}, c-DPEA \cite{ming2021dual}, CCMODE \cite{wang2018cooperative}, CMME \cite{ming2022constrained}, NSGA-II-CDP \cite{deb2002fast} are current popular CMOEAs to solve the CMOPs. For fairness, the shared parameters (i.e., the population size and the maximal number of function evaluations) of all CMOEAs were the same as those of our LSCV-MCEA. Meanwhile, all CMOEAs adopted the parameter settings in the original papers. The parameter settings are described in the additional file 1. In addition, these CMOEAs are under the framework of MDS, NCUA and DFVS in this work and denoted as special forms. For example, LSCV-MCEA under the framework of MDS, NCUA and DFVS are denoted as LSCV-MCEA\_MDS, LSCV-MCEA\_NCUA and LSCV-MCEA\_DFVS, respectively. Other constrained multi-objective evolutionary algorithms are all same as above.

ii) Network and differential expression genes (DEG)-based methods. The network-based methods for identifying PDTs were also taken for comparison in this paper, including CPGD \cite{guo2021network}, PNC \cite{guo2019novel}, ActiveDriver \cite{reimand2013systematic}, OncoDriveFM \cite{gonzalez2012functional}, DriverML \cite{han2019driverml} and Hub-genes. The DEG-based methods consist of DEG-Folchange, DEG-p-value, and DEF-FDR. All the above methods were executed on the same TCGA datasets (i.e., BRCA, LUSC, and LUAD) as our LSCV-MCEA according to their manuals.The details for the above methods were shown in Section S-IV of additional file 1.

iii) Traditional structural network control methods, including MMS, MDS, NCUA, and DFVS.

\subsubsection{Performence metrics} 
To validate the effectiveness of LSCV-MCEA on the three cancer datasets (i.e., BRCA, LUAD, and LUSC), HV \cite{zitzler1999multiobjective}, IGD \cite{bosman2003balance}, and AUC were adopted to evaluate its performance in identifying PDGs for each PGIN. Meanwhile, to evaluate the usage efficiency of PDTs for personalized drug discovery of different methods, the number of PDTs matching with drug combinations was calculated, and the anti-cancer drug combinations for each patient were ranked. The drug combinations annotated in the CAC drugs were applied to obtain the AUC \cite{lobo2008auc} of the top-ranked/predicted anti-cancer drug combinations from different methods. The AUC value of the predicted anti-cancer drug combinations was obtained based on the predicted probability and the true label in the CAC. The computational details for obtaining the probability was shown in Section S-V of additional file 1.

The inverted generational distance (IGD) and hypervolume (HV) parameters were exploited to reflect the algorithm convergence and diversity to find multiple sets of drug targets when solving the MONCP in PGIN. Two parameters, namely the true Pareto front (PF) and the reference point for obtaining the HV and IGD, are major determinants.  For the IGD indicator, it needs to calculate the distance between the solutions of the algorithm and the true Pareto front to evaluate the convergence and diversity of the algorithm. However, for our DMOM-LSCV it is difficult to know the true Pareto front and the exhaustive method does not work well due to the large scale of decision variables. Therefore, we took the union set of the pareto fronts from different algorithms, and selected the non-dominated solutions of this union set as the true Pareto front to ensure better diversity distribution of optimal solutions in the objective space. In addition, this strategy has also been widely adopted in previous studies \cite{yang2021multi,he2020paired}. To calculate the HV, the objective values were normalized, and (1.1,1.1) was used as the reference point in the normalized objective space. For different CMOEAs, the values of HV and IGD under 30 independent runs were recorded. In this work, the Wilcoxon rank-sum test \cite{derrac2011practical} at a significance level of 0.05 was adopted to perform statistical analysis.

\begin{figure*}[!t]
	\centering
	\includegraphics[scale=0.58]{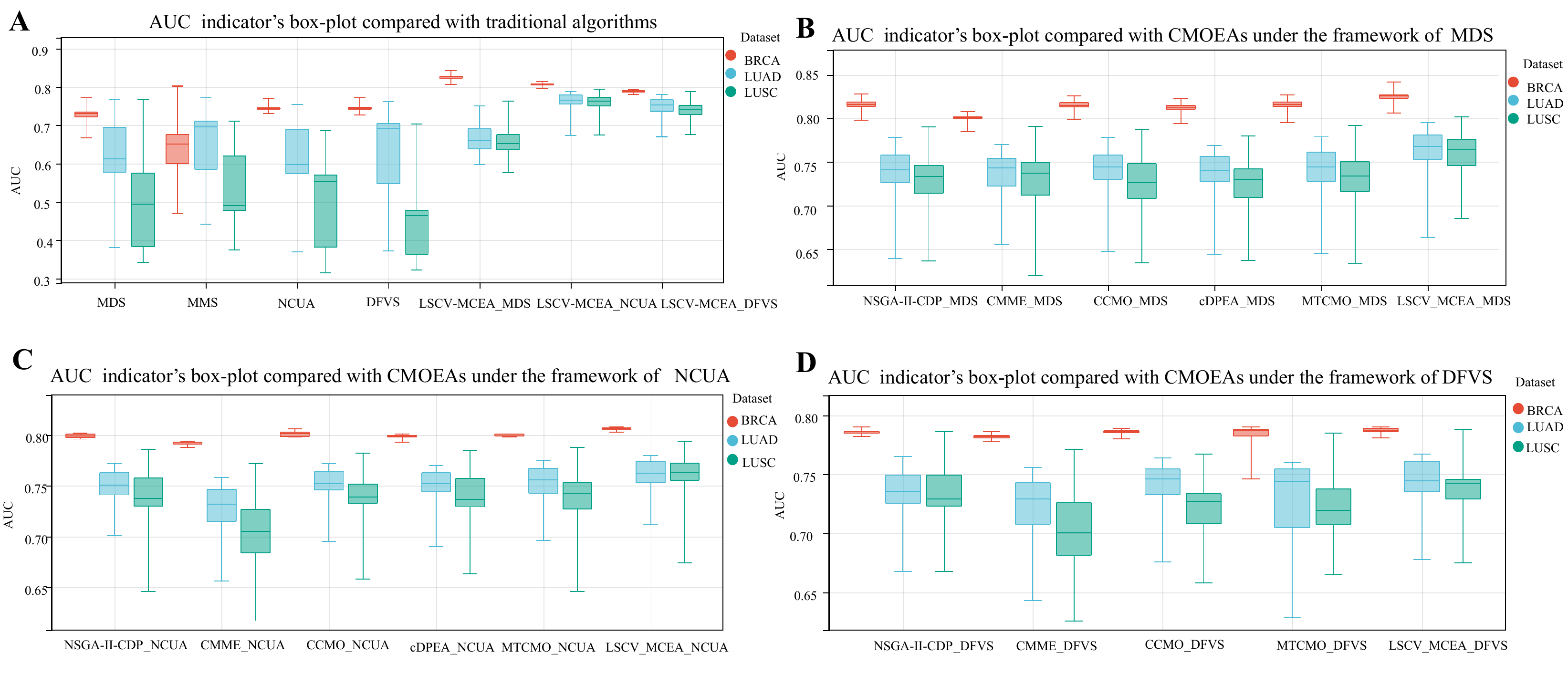}
	\caption{{\textbf{The comparison results of AUC indicator for different methods.} (A) The comparison results of AUC indicator for LSCV-MCEA and traditional single objective based network control methods (i.e.,MDS, DFVS and NCUA). (B-D) The comparison results of AUC indicator for LSCV-MCEA and other CMOEAs under the framework of (B) MDS, (C) NCUA and (D) DFVS.  
	}}
	\label{fig3}
\end{figure*}

\subsection{Performance of LSCV-MCEA for discovering clinical combinatorial drugs}

As described above, LSCV-MCEA was designed by considering the MONCP, which requires a control on the system with minimum driver nodes and needs to find maximum prior-known drug-target genes. To show the effectiveness of MONCP, compared with the single optimization objective-based methods including the traditional structural network control methods (i.e., MDS, MMS, NCUA, and DFVS) and other driver genes (candidate drug targets) methods (i.e., Network and differential expression gene (DEG)-based methods), their AUC indicator was calculated on three cancer datasets. The greater the AUC of a given method, the better the performance in discovering clinical combinatorial drugs. 

\subsubsection{Advantage of LSCV-MCEA over traditional structural network control methods} 
Our LSCV-MCEA outputs MCDS, from which the genes whose frequency is larger than 0.8 were selected as the driver genes (candidate drug targets) for an individual patient. For traditional structural network control-based methods (i.e., MDS, MMS, DFVS, and NCUA), one optimal set of driver nodes with the minimum number could be obtained on the PGIN of an individual patient. Then, the ranking of the combinatorial drugs could be obtained by targeting the driver genes and assigning a probability to each combinatorial drug for an individual patient (see was shown of additional file 1). Based on the predicted probability and the true label in the CAC, the AUC value of the predicted anti-cancer drug combinations can be acquired for an individual patient. The comparison results of LSCV-MCEA and the traditional structural network control methods are illustrated in \textbf{Figure 3A}. It can be seen from this figure that LSCV-MCEA achieved a higher AUC than traditional structural network control methods on these three cancer datasets. These results demonstrate that our LSCV-MCEA could more effectively detect drug targets for discovering anti-cancer drug combinations than traditional structural network control methods.

\subsubsection{Advantage of LSCV-MCEA over other CMOEAs} 
To further show the advantage of LSCV-MCEA for discovering clinical combinatorial drugs, the AUC of LSCV-MCEA was compared with that of other six CMOEAs for each patient, and these methods all output MCDS  in a PGIN. To be fair, for all CMOEAs, the best AUC value among the PS of an individual patient was obtained. \textbf{Figures 3(B-D)} show the box plots of the AUC indicator of LSCV-MCEA and other six CMOEAs for individual patients on the three cancer datasets under the framework of MDS, NCUA, and DFVS, respectively. Note that if the results of the algorithms are not shown in the box plot, it is indicated that no feasible solutions are found, and the AUC cannot be calculated for the corresponding algorithm. The results of \textbf{Figures 3(B-D)} indicate that higher AUC can be obtained on the PS of LSCV-MCEA, compared with those of the other six CMOEAs. The results demonstrate that LSCV-MCEA had better performance than other CMOEAs in discovering clinical combinatorial drugs on these cancer datasets. The detailed results of all patients in \textbf{Figure 3} were provided in additional file 2.

\begin{figure}[!t]
	\centering
	\includegraphics[scale=0.43]{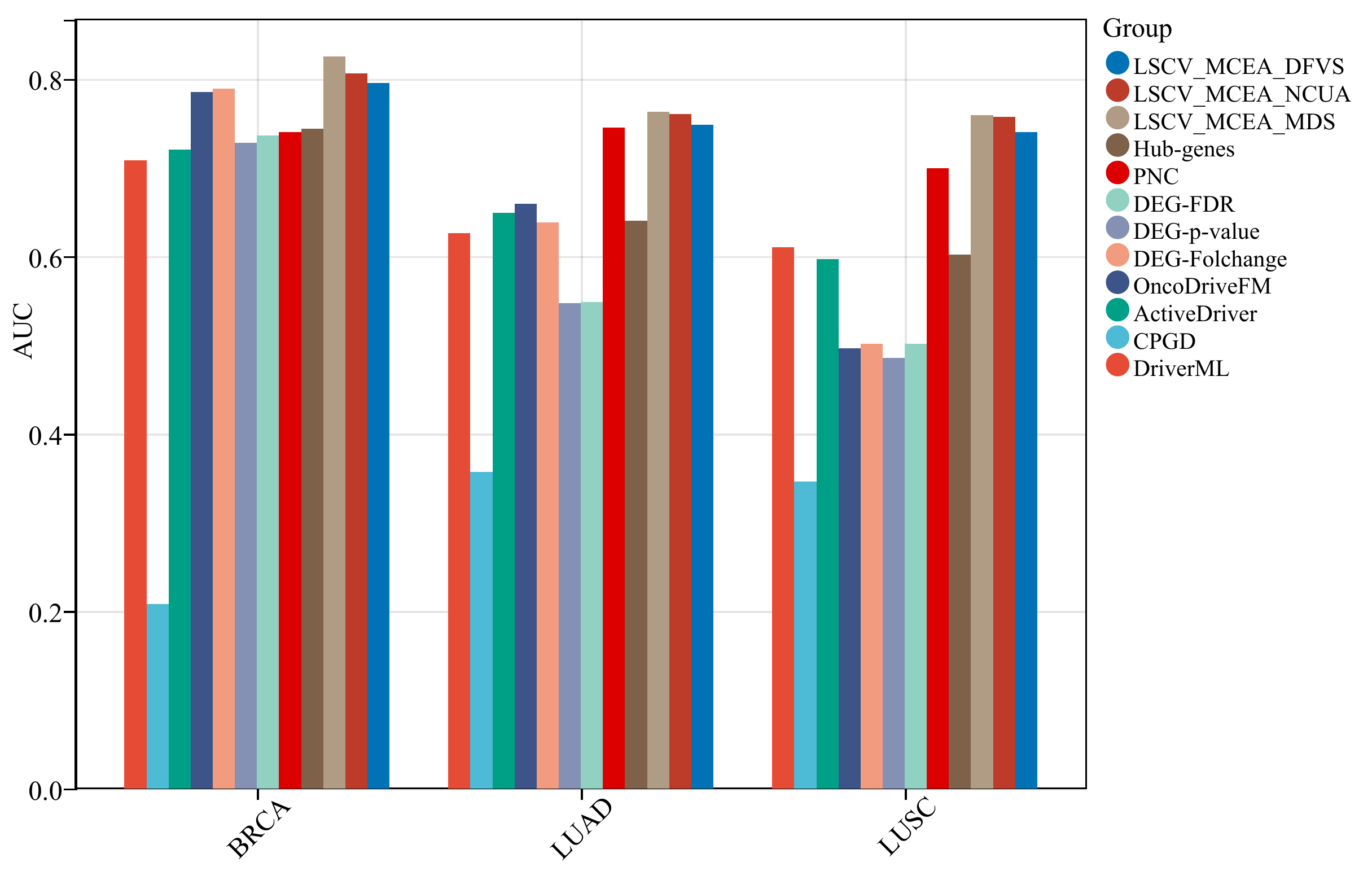}
	\caption{{\textbf{The comparison results of AUC indicator for LSCV-MCEA and other cancer specific driver genes based methods on three cancer datasets}}}
	\label{fig4}
\end{figure}

\begin{figure*}[!t]
	\centering
	\includegraphics[scale=1.05]{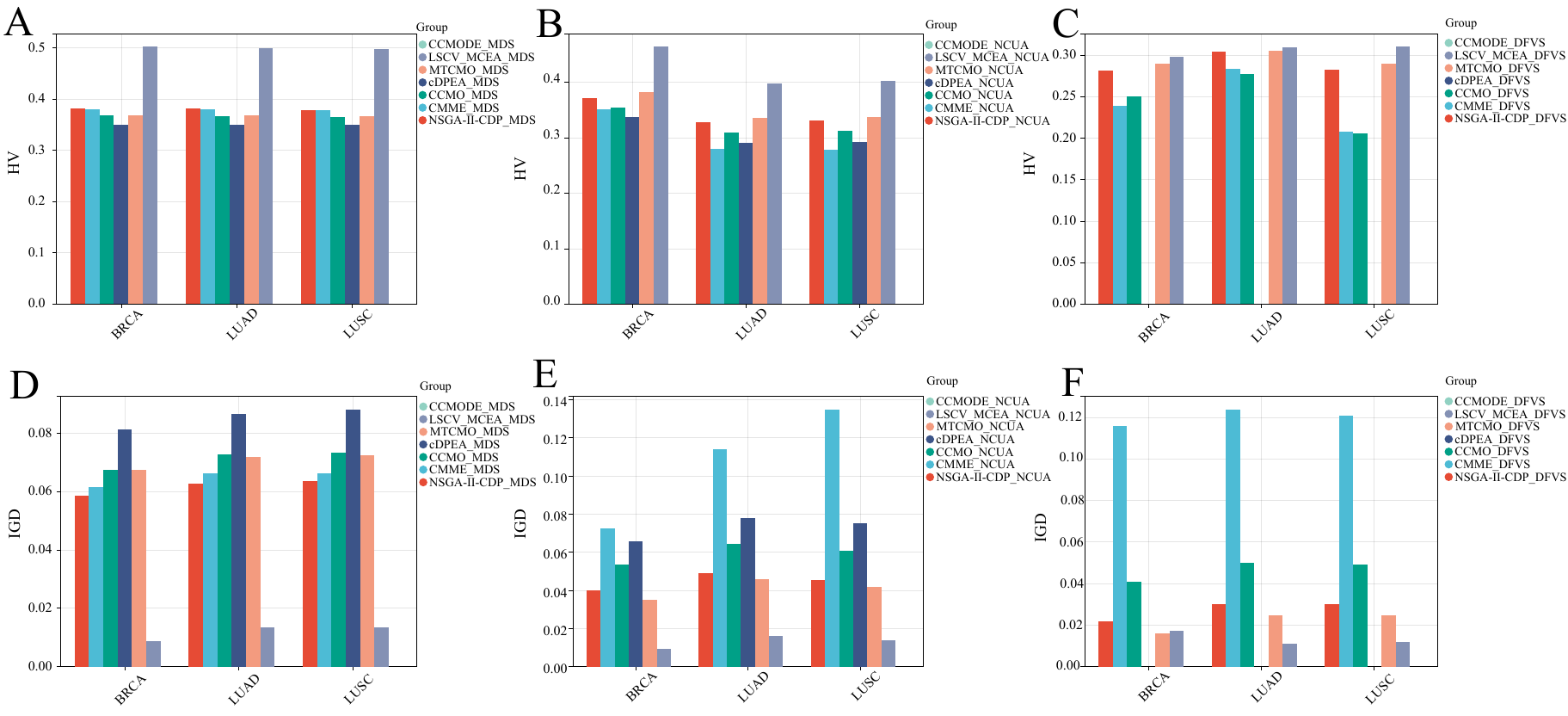}
	\caption{\textbf{The comparison results of HV and IGD indicator for seven CMOEAs on three cancer datasets.} \textbf{(A-C)} HV indicator of seven CMOEAs on three cancer datasets under the framework of (A) MDS, (B) NCUA and (C) DFVS. \textbf{(D-F)} IGD indicator of seven CMOEAs on three cancer datasets under the framework of (D) MDS, (E) NCUA and (F) DFVS.}
	\label{fig5}
\end{figure*}

\subsubsection{Advantage of LSCV-MCEA over other methods based on cancer-specific driver genes} 
To show the advantage of LSCV-MCEA over other methods based on cancer-specific driver genes, our LSCV-MCEA obtained cancer-specific driver genes by integrating the driver genes of all individual patients on the three cancer datasets. First, for a given cancer dataset, the gene frequency appearing on the PS was calculated, and the genes whose frequency was larger than 0.8 were selected as the candidate PDGs for individual patients. Then, the genes whose frequency was larger than 0.8 among all of the candidate PDGs were considered, and the corresponding cancer-specific driver genes were obtained on the three cancer datasets. Finally, the anti-cancer drug combinations were ranked by calculating the number of cancer-specific driver genes matching with drug combinations. Note that a probability was assigned to each combinatorial drug for an individual patient according to its rank (see section S-V of additional file 1), and the AUC values of the predicted anti-cancer drug combinations were obtained on the three cancer datasets. The experimental results are presented in \textbf{Figure 4}. According to these results, LSCV-MCEA outperformed other methods on these three cancer datasets. For example, as shown in \textbf{Figure 4}, the AUC values of LSCV-MCEA for MDS, DFVS, and NCUA on BRCA were 0.825, 0.787, and 0.807, respectively, which are much larger than those of the other methods. These results demonstrate that our LSCV-MCEA can more effectively identify cancer driver genes to discover clinical combinatorial drugs than other advanced methods based on cancer-specific driver genes on these cancer datasets. The detailed results of all patients in \textbf{Figure 4} were provided in additional file 3.
	
\subsection{Performance of LSCV-MCEA in terms of algorithm convergence and diversity for solving MONCP}
To show the advantages of LSCV-MCEA in terms of algorithm convergence and diversity over other advanced CMOEAs, HV and IGD were used to evaluate the performance of different methods for finding multiple sets of drug targets when solving MONCP in the PGIN. A smaller IGD value and a larger HV value indicate better performance of the algorithm. Here, the average value of HV and IGD under 30 independent runs was recorded. \textbf{Figure 5} shows the mean HV and IGD values, which were obtained by seven CMOEAs on the three cancer datasets, respectively. From these results, it can be seen that the proposed LSCV-MCEA obtained larger HV and smaller IGD values than the other six CMOEAs on the three cancer datasets. These results demonstrate that our LSCV-MCEA performed better than the other six CMOEAs for finding multiple sets of drug targets when solving MONCP in terms of algorithm convergence and diversity on the three cancer datasets. The detailed results of all patients in \textbf{Figure 5} were provided in additional file 4.

\begin{figure*}[!t]
	\centering
	\includegraphics[scale=1.05]{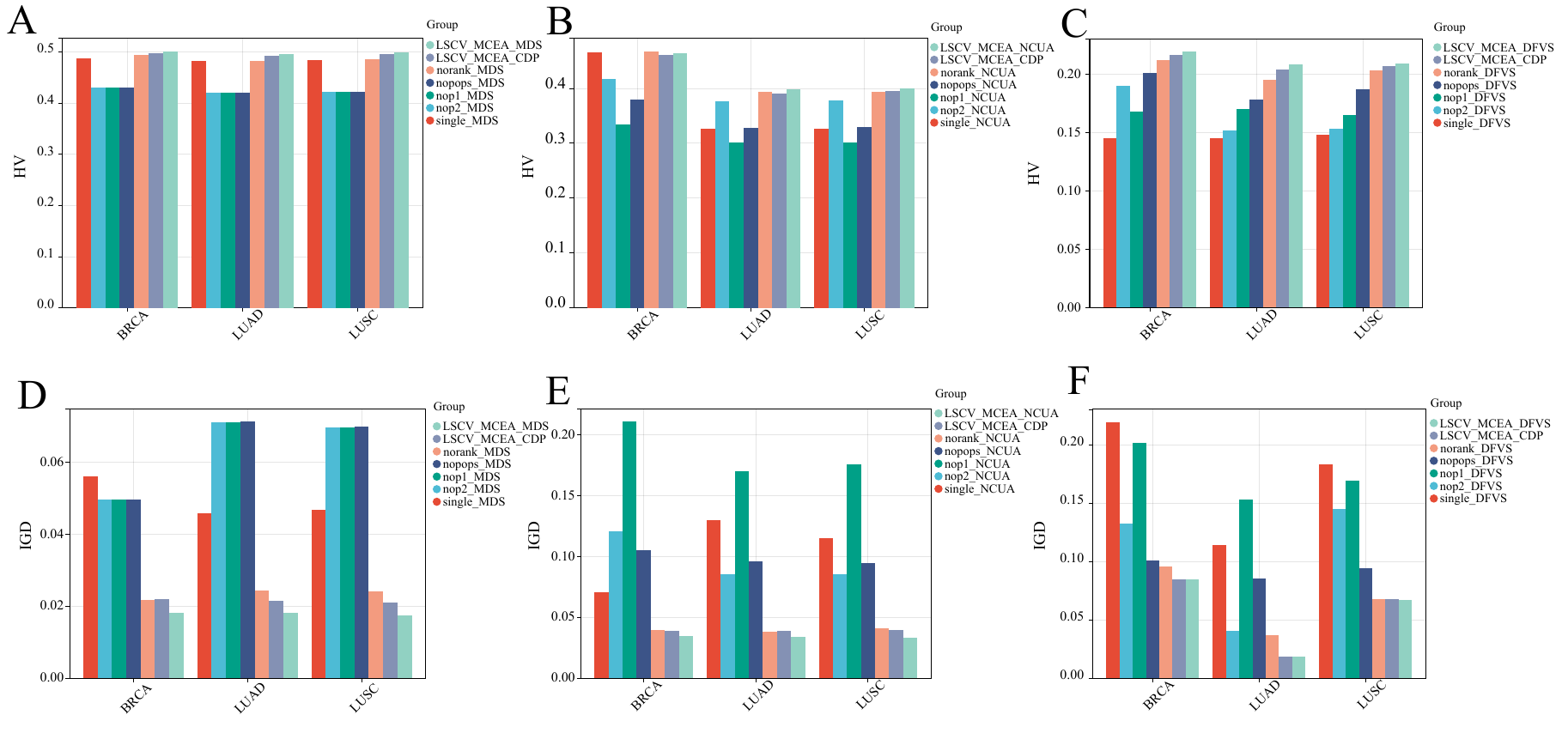}
	\caption{{\textbf{The comparison results of HV and IGD indicator for LSCV-MCEA and other variants on three cancer datasets.} \textbf{(A-C)} HV indicator of LSCV-MCEA and other variants on three cancer datasets under the framework of (A) MDS, (B) NCUA and (C) DFVS. \textbf{(D-F)} IGD indicator of LSCV-MCEA and other variants on three cancer datasets under the framework of (D) MDS, (E) NCUA and (F) DFVS. }}
	\label{fig6}
\end{figure*}

\hspace*{\fill} 

\begin{figure}[!t]
	\centering 
	\includegraphics[scale=1]{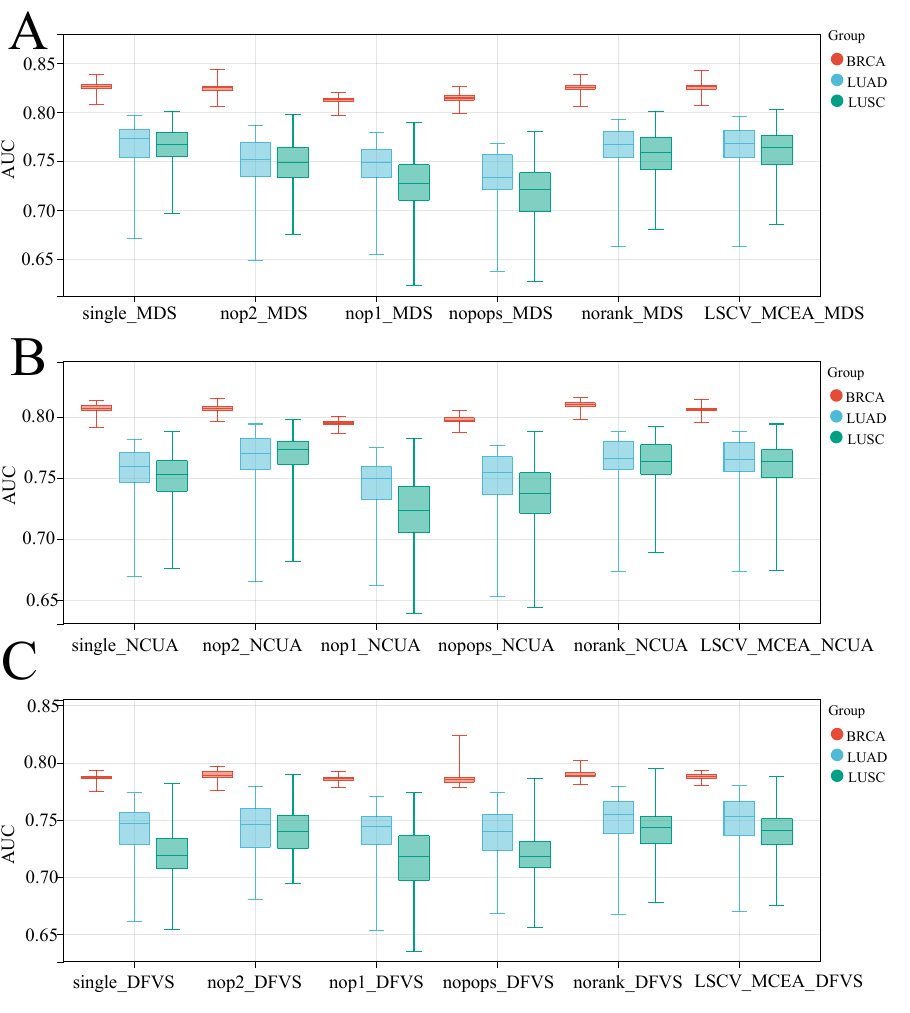}
	\caption{{\textbf{The comparison results of AUC indicator for LSCV-MCEA and other variants on three cancer datasets under the framework of (A) MDS, (B) NCUA and (C) DFVS.}}}
	\label{fig7}
\end{figure}

\subsection{Effectiveness of the multi-tasking framework and the rankings-based fitness function in LSCV-MCEA}
The advantage of our LSCV-MCEA is adopting specific strategies for handling large-scale CMOPs for drug target identification. In our LSCV-MCEA there are two main components in the algorithm, i.e. the multi-tasking strategy and the constraint handling techniques (CDP) ($\varepsilon$ constrained method on main population and rankings-based fitness function strategy on two subpopulations). To verify the effectiveness of the operators in our LSCV-MCEA,  LSCV-MCEA with and without these operators was analyzed. Specifically, the HV and IGD of the following variants were calculated: LSCV-MCEA without two sub-populations, LSCV-MCEA without sub-population1, and LSCV-MCEA without sub-population2, denoted as ${nopops}$, $ {nop1}$, and ${nop2}$, respectively. As shown in \textbf{Figure 6}, the proposed LSCV-MCEA had a larger HV and a smaller IGD on the three cancer datasets compared with LSCV-MCEA without the multi-tasking framework (i.e., LSCV-MCEA without sub-populations and LSCV-MCEA with either of the two sub-populations) under the framework of MDS, NCUA, and DFVS. Additionally, the performance of the original LSCV-MCEA was better (a larger HV and a smaller IGD) than the LSCV-MCEA without the rankings-based fitness function (denoted as $norank$). In addition, the performance of different constraint handling techniques (CDP) and single learning between the populations on the algorithm performance was discussed, denoted as $LSCV-MCEA\_CDP$ and $single$. These rsesults indicate that the multi-tasking framework and the rankings-based fitness function can effectively improve the performance of LSCV-MCEA. The detailed results of all patients in \textbf{Figure 6} were provided in additional file 5.

Furthermore, this paper compared the AUC values of LSCV-MCEA and those of other variants for discovering clinical combinatorial drugs. The rsesults are presented in \textbf{Figure 7}. It can be seen that the multi-tasking framework with sub-populations and the rankings-based fitness function enable LSCV-MCEA to better discover clinical combinatorial drugs on these cancer datasets, especially on the LUSC and LUAD cancer datasets.The detailed results of all patients in \textbf{Figure 7} were provided in additional file 6.

\subsection{Biological significance of the LSCV-MCEA model}
This section mainly discusses the biological significance of the multiple sets of drug targets (i.e., PS) in our LSCV-MCEA from the following two aspects. 

First it was validated whether the PS of our LSCV-MCEA was significantly enriched in the Cancer Gene Census (CGC) dataset as well-established driver gene sets \cite{futreal2004census}. To investigate whether the solutions (i.e., the set of driver genes) in the PF are enriched in the given CGC dataset, this paper calculated the empirical p-value  for each solution in the PS enriching in the CGC dataset on the three cancer datasets under the framework of MDS, NCUA, and DFVS. The computational details of obtaining p-value were shown in section S-VI of additional file 1. The fraction of solutions whose p-values are smaller than 0.05 was considered as the Enrichment Significance Score (ESS) for a given PS. The results are shown in \textbf{Figure 8}. From \textbf{Figure 8A}, several observations are obtained:

\begin{figure*}[!t]
	\centering
	\includegraphics[scale=0.4]{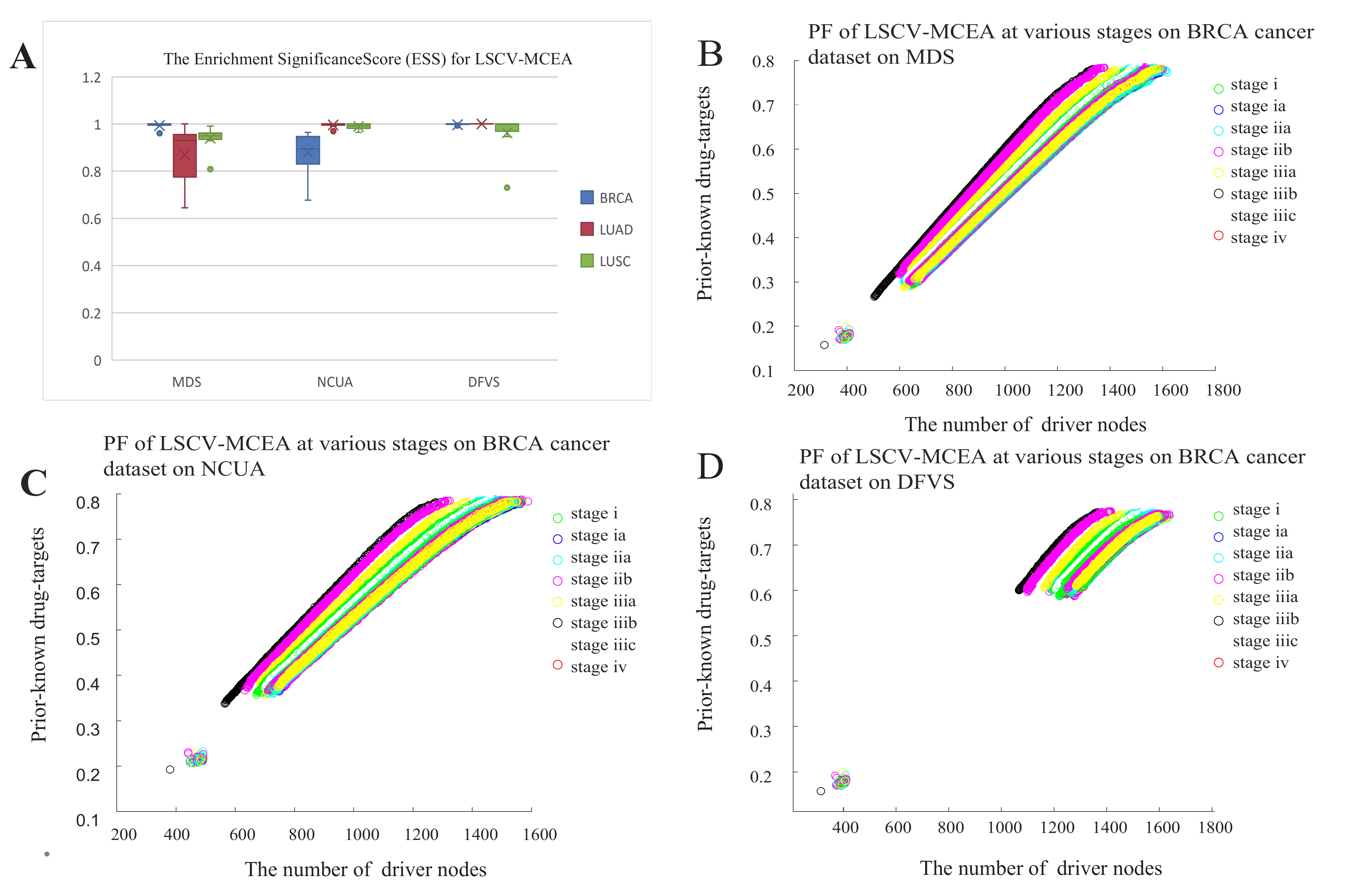}
	\caption{{\textbf{Verification of biological significance.} \textbf{(A)} The result of the set of driver genes  enriched in the given CGC dataset for LSCV-MCEA. \textbf{(B-D)} PF of LSCV-MCEA for individual patients at various stages on BRCA cancer dataset under the framework of (B) MDS, (C) NCUA and (D) DFVS. }}
	\label{fig8}
\end{figure*}

i) The ESS score was generally larger than 0.8. This indicates that there were at least $80\%$  efficient solutions on the PS, which were significantly enriched in the CGC for LSCV-MCEA under MMS, NCUA, and DFVS on the three cancer datasets. These results demonstrate that most of the solutions in the PS of LSCV-MCEA contribute to the discovery of cancer-driver genes. 

ii) The performance of LSCV-MCEA varied under different network control frameworks. Overall, LSCV-MCEA with the DFVS control framework performed better than NCUA and MDS control frameworks.

Then, whether the PF of cancer patients could detect significant disease signals during cancer progress was investigated. \textbf{Figures 8(B-D)} present all the PFs of LSCV-MCEA under MDS, DFVS, and NCUA control frameworks for individual patients in various stages on the BRCA cancer dataset. Also, the values of the two objectives in our MONCP are provided for the solutions of the traditional network control frameworks (i.e., MDS, DFVS, and NCUA). It can be seen from the results that the PF of MDS, DFVS, and NCUA could efficiently detect the disease signals of stage iiib compared with the traditional network control frameworks. Meanwhile, the solutions in the PF of our LSCV-MCEA were not dominated by the traditional network control framework, indicating that our LSCV-MCEA complemented the traditional network control principles for describing network control characteristics for BRCA cancer patients.

To further demonstrate the complementary relationship between the solutions of our LSCV-MCEA and the traditional network control principles, this paper considered the differential genes between our LSCV-MCEA and the traditional network control principles for individual patients on the three cancer datasets. The differential genes are those in the PS of LSCV-MCEA but not in the set of driver genes for the traditional network control principles. The differential genes were ranked by calculating their mean frequency appearing in the PS for all individual patients, and the top 50 ranking genes were selected as the candidate drug targets fort the subsequent analysis.The gene list of top 50 ranking genes for LSCV-MCEA under framework of MDS, DFVS and NCUA in three cancer datatsets was provided in additional file 7.

Finally, through searching for drug target information on drug response datasets (i.e., the Genomics of Drug Sensitivity in Cancer (GDSC) dataset \cite{yang2012genomics} and the iGMDR database \cite{chen2020igmdr}), it was validated whether the top 50 ranking genes could provide efficient drug information for individual patients on the three cancer datasets. The statistic information of LSCV-MCEA with MDS, DFVS, and NCUA on the three cancer datasets is presented in Table S1 of additional file 1. It can be seen from Table S1 that MDS, DFVS, and NCUA could find different drug targets for discovering efficient drugs on the BRCA cancer dataset. These results demonstrate that different network control principles are complementary and could provide different novel drug targets by using our LSCV-MCEA. More details of the drug target anofd the corresponding drug information are provided in additional file 1.

In addition, picking few sets for visualization and gene/pathway enrichment analysis for molecular medicine insights with p-values is necessary. First,  genes that appeared in the identified Pareto optimal solution set (PS) with a frequency equal to 1 in every BRCA patient by expecting to obtain the overlapping genes in PS were selected. These selected genes were considered as candidate personalized driver genes. Then these personalized driver genes of patients at the same stage were integrated to obtain the stage specific driver genes of different stages during cancer progress. Finally, through pathway enrichment analysis of KEGG and WikiPathway for these stage specific driver genes of different stages, the results of the stage specific pathways with the top 20 significant p-values are presented in Figures S4-S5 in additional file 1.  It can be seen that our LSCV-MCEA could discover certain cancer pathways that are related to patients with relatively early, intermediate and advanced stages of BRCA cancer.

\section{Conclusion}
In this study, the concept of multi-objective optimization based structural network control principles (called MONCP) under the framework of DFVS, MDS and NCUA was proposed by considering minimum driver nodes and maximum prior-known drug-target information. Different from traditional structural network control principles, our MONCP provides not only  multiple sets of drug targets but also more solutions with more drug target information. Then a discrete multi-objective optimization problem with large-scale constrained variables was built and we devoloped a novel multi-objective evolutionary optimization model (denoted LSCV-MCEA) to solve MONCP. In LSCV-MCEA, multi-tasking framework and rankings-based fitness method were used under the framework of CMOEA on the PGIN. Using breast and lung cancer datasets, we validated that compared with other advanced methods, LSCV-MCEA can more effectively identify the PDTs with the highest Area Under The Curve (AUC) score for predicting clinical annoted combinatorial drugs. Compared with other existing evolutionary optimization methods, LSCV-MCEA had better convergence and diversity of algorithms in terms of HV and IGD for solving MONCP on three cancer datasets. Besides, we validated that  the multi-tasking framework and the rankings-based method can effectively improve the performance of the LSCV-MCEA algorithm in terms of HV, IGD, and AUC indicators for solving MONCP. 

In addition, we validated that the PS of our LSCV-MCEA were significantly enriched in CGC dataset on these three cancer datasets, in which some novel drug targets (i.e., AKT1 and BRAF for NCUA, AR and PTPN6 for DFVS and AKT1, KIT and SMARCA2 for MDS on BRCA cancer datasets) with corresponding approved cancer type specific drugs via the SurvExpress tool \cite{aguirre2013survexpress}, iGMDR dataset \cite{chen2020igmdr}. Another interesting finding is that our LSCV-MCEA on three frameworks could efficiently detect the disease signals of stage iiib for BRCA cancer. Then, we also did pathway enrichment analysis of KEGG and WikiPathway and find that the redundancies among multiple driver node sets along the EMO pareto front existed.

In the future, we can propose more effective strategies (i.e., sparse optimization strategies of CMOPs \cite{wang2021enhanced} to improve the performance of identifying drug targets in the light of MONCP whose variables suffer from a large-scale sparse constraints in decision space. In addition, it will be another promising direction by considering multi-modal optimization on MONCP to find more effective solutions in PGIN \cite{liang2022multiobjective}.

\bibliographystyle{unsrt}
\bibliography{mybibfile}

\end{document}